\documentclass[10pt,twocolumn,letterpaper]{article}

\usepackage[pagenumbers]{cvpr} 

\usepackage[accsupp]{axessibility}

\usepackage[dvipsnames]{xcolor}

\usepackage{floatrow}
\usepackage{braket}
\usepackage{nicefrac}
\usepackage{array}
\usepackage{multirow}
\usepackage{placeins}
 \usepackage{mathtools}
\usepackage{rotating}
\usepackage{comment}
\usepackage{algorithm}
\usepackage{algpseudocode}
\algnewcommand{\LeftComment}[1]{\Statex \(\triangleright\) #1}
\usepackage{subcaption}
\usepackage[export]{adjustbox}
\usepackage{booktabs}

\newcommand{\vct}[1]{{#1}}
\DeclareMathOperator*{\argmax}{arg\,max}
\DeclareMathOperator*{\argmin}{arg\,min}
\newcommand{\RNum}[1]{\uppercase\expandafter{\romannumeral #1\relax}}

\renewcommand{\paragraph}[1]{{\vspace{1mm}\noindent \bf #1}}

\newcommand{\XSet}{\mathbb{Z}^N_2}

\definecolor{cvprblue}{rgb}{0.21,0.49,0.74}
\usepackage[pagebackref,breaklinks,colorlinks,citecolor=cvprblue]{hyperref}

\def\paperID{xxx} 
\def\confName{CVPR}
\def\confYear{2024}

\title{Probabilistic Sampling of Balanced K-Means\\using Adiabatic Quantum Computing}

\author{\begin{tabular}{ccccccccccccccc}\multicolumn{3}{c}{Jan-Nico Zaech \textsuperscript{1,2}} & \multicolumn{3}{c}{Martin Danelljan \textsuperscript{1}} & \multicolumn{3}{c}{Tolga Birdal \textsuperscript{3}} & \multicolumn{3}{c}{Luc Van Gool \textsuperscript{1,2}}\end{tabular}\\\\
{\small \textsuperscript{1}ETH Zurich, Switzerland \qquad \textsuperscript{2} INSAIT, Sofia University, Bulgaria}\\
{\small\textsuperscript{3}Imperial College London, United Kingdom}}

\begin{document}
\maketitle
\begin{abstract}
Adiabatic quantum computing (AQC) is a promising approach for discrete and often NP-hard optimization problems. Current AQCs allow to implement problems of research interest, which has sparked the development of quantum representations for many computer vision tasks. Despite requiring multiple measurements from the noisy AQC, current approaches only utilize the best measurement, discarding information contained in the remaining ones. In this work, we explore the potential of using this information for probabilistic balanced k-means clustering. Instead of discarding non-optimal solutions, we propose to use them to compute calibrated posterior probabilities with little additional compute cost. This allows us to identify ambiguous solutions and data points, which we demonstrate on a D-Wave AQC on synthetic tasks and real visual data.
\end{abstract}

\FloatBarrier    
\section{Introduction}
\label{sec:intro}

\begin{figure}[t]
    \centering
    \captionsetup{width=\linewidth}
    \includegraphics[width=\linewidth]{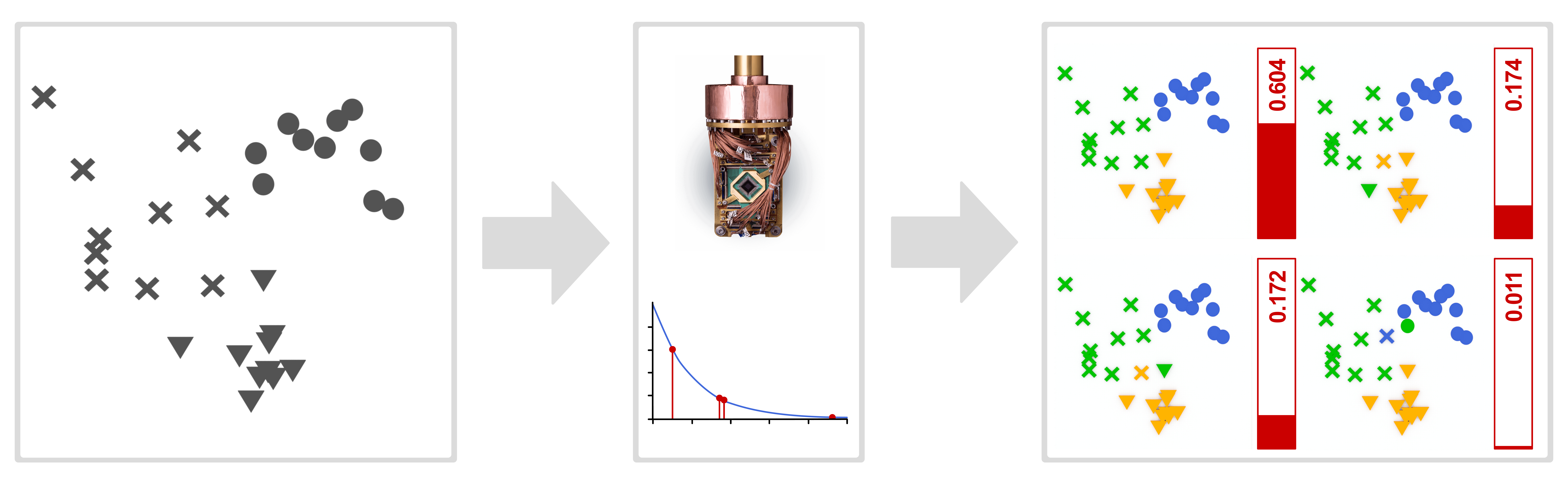}
    \caption{The proposed approach uses an adiabatic quantum computer to sample solutions of a balanced k-means problem. By using an energy-based formulation, likely solutions are drawn from a Boltzmann distribution. By reparametrizing the distribution, the calibrated posterior probability of each solution can be estimated.\vspace{-3mm}}
    \label{fig:title}
\end{figure}

Clustering and uncertainty quantification (UQ) are fundamental problems in machine learning and computer vision. Clustering, the task of grouping objects based on the similarity of their features, plays a pivotal role in analysis and organization of vast quantities of unlabeled visual data with applications including image classification \citep{lazebnik_bags_2006,caron_deep_2018, nguyen_quantum_2023}, segmentation \citep{coleman_image_1979,dhanachandra_image_2015}, tracking \citep{keuper_motion_2020}, and network training \citep{xu_survey_2005,coates_learning_2012,yang_joint_2016}. UQ, on the other hand, aims to assess the trustworthiness of predictions and accounts for the impact of data variability. As a crucial element for certifying confidence in results, UQ is indispensable in clustering, especially given the often ambiguous nature of data partitioning~\citep{he2023survey}.

Clustering typically involves solving for an optimal assignment of data points to (latent) centroids, while a full Bayesian UQ approach strives to accurately depict the posterior distribution. Unfortunately, both of these problems are known to be notoriously difficult, classified within the NP complexity class\footnote{Computing marginal probabilities on a Bayesian network is, basically, a counting problem that can become arbitrarily hard to solve.}~\cite{cooper1990computational,dasgupta2008hardness,mahajan2012planar}. These challenges within the Turingian modes of computation have motivated: (i) continuous relaxations in clustering~\cite{zha2001spectral} and (ii) the advent of approximate Bayesian inference methods for UQ, including the EM algorithm \cite{dempster_maximum_1977}, variational inference \cite{bishop2006pattern}, and sequential MCMC \cite{del2006sequential}.

In this paper, departing from the above-mentioned approximations or continuous relaxations, we take a completely different approach and harness the upcoming and novel computational paradigm of quantum computers to tackle the challenging task of clustering with \emph{calibrated} uncertainty quantification. Specifically, we focus on the widely used balanced K-means~\cite{lloyd1982least,macqueen1967some,malinen_balanced_2014}, which is an iterative algorithm that first assigns each data point to a cluster followed by an update of cluster centroids. 

Despite their early state, quantum machine learning algorithms have approached tasks such as optimization of quadratic problems~\citep{kadowaki_quantum_1998}, training of restricted Boltzmann machines~\citep{dixit_training_2021}, and learning with quantum neural networks~\citep{abbas_power_2021}. While the current quantum computers can only solve small-scale problems, they provide the basis to develop and test algorithms that can considerably increase the size of feasible problems in the future.
In our work, we interpret K-means clustering in an energy-based framework and aim to model the posterior distribution by sampling multiple high-probability binary assignments from it on actual quantum hardware at little additional cost. We specifically use the \emph{quantum annealer} (QA) of D-Wave~\cite{mcgeoch_dwave_2022}, which follows the concept of \emph{adiabatic quantum computing} (AQC)~\cite{biamonte_realizable_2008} and avoids simulating costly Markov chains as in classical computers. 
To achieve this, we first formulate the k-means objective as a quadratic energy function over binary variables and embed the clustering task into the quantum-physical system of D-Wave. By repeatedly measuring the quantum system, we sample high-probability solutions according to the Boltzmann distribution. Unlike previous approaches that only use the best solution and discard all other measurements, we utilize all samples to generate probabilistic solutions for the k-means problem, as shown in Figure~\ref{fig:title}. Due to temperature mismatch~\citep{pochart_challenges_2021}, AQCs sample from a modified posterior instead of the true posterior, which needs to be \emph{re-calibrated} for the task at hand. We do so by estimating the posterior probability for each solution. This allows us to identify ambiguous points and provide alternative solutions. 
By leveraging {QA}, our approach respects the binary nature of the assignment problem.

We demonstrate our algorithm on a D-Wave quantum computer and also perform extensive experiments in simulation. On real data, we show that our approach can be utilized for distributions that do not strictly follow the initial assumptions and that it is suitable for identifying ambiguous images. 
Our primary contributions are:
\begin{itemize}
    \setlength{\itemsep}{0pt}
    \setlength{\parskip}{0pt}
    \item A quantum computing formulation of balanced k-means clustering that predicts calibrated confidence values and provides a set of alternative clustering solutions.
    \item A reparametrization approach used to calibrate posterior probabilities from samples that avoids exact tuning of the AQC sampling temperature.
    \item Extensive experiments on synthetic and real data showing the calibration of our approach both on simulations and on the D-Wave Advantage 2 QA prototype.
\end{itemize}
Developing better heuristics for the NP-hard challenges of clustering and UQ presents significant difficulties for classical approaches. However, we are optimistic that the ongoing progress in quantum computing technology will progressively enhance the effectiveness of our algorithms.
\section{Related Work}
\paragraph{Quantum computation.}
With the availability of quantum computers to the general research community~\citep{bunyk_architectural_2014, mcgeoch_advantage_2021, mcgeoch_dwave_2022,debnath_demonstration_2016}, the research interest in finding applications for such systems has considerably increased. In this context AQC~\citep{kadowaki_quantum_1998, bunyk_architectural_2014} provides a well-tangible starting point, even though many applications need a complete reformulation considering the architectural differences of a quantum computer.
Current applications for AQC are optimization tasks in different fields~\citep{neukart_traffic_2017, ohzeki_control_2019, mulligan_designing_2020, mugel_hybrid_2021}, including \emph{quantum computer vision}~\cite{birdal_quantum_2021}, where a strong interest in finding quantum computing formulations has developed. These are often related to hard permutation problems~\cite{benkner_adiabatic_2020, birdal_quantum_2021, benkner_qmatch_2021} like tracking~\citep{zaech_adiabatic_2022}, graph, shape and point matching~\cite{benkner_qmatch_2021, bhatia_ccuantumm_2023, meli_iterative_2022} or training binary (graph) neural networks~\cite{krahn2023projected}. In this context, \citet{birdal_quantum_2021} evaluate the k-best solutions, however, without the energy-based formulation, only little improvement is achieved.
Another tasks of interest for the community is model fitting to find camera parameters~\cite{doan_hybrid_2022} or separating motion components~\cite{arrigoni_quantum_2022}, which can be formulated as a quantum-computing problem~\cite{wright_benchmarking_2019, chin_quantum_2021, arrigoni_quantum_2022, doan_hybrid_2022, chi_programmable_2022, farina_quantum_2023} instead of consensus maximization.

\paragraph{Clustering.}
Clustering is a well-studied problem for quantum- as well as quantum-inspired algorithms~\citep{aimeur_quantum_2007}. It groups a dataset $\vct X$ into disjoint clusters ${c_1, ..., c_K}$, with each cluster containing points similar in the feature space. For example, k-means algorithm~\citep{forgy_cluster_1965, lloyd_least_1982} utilizes quadratic distances, favoring compactness, whereas dbscan~\citep{ester_densitybased_1996} identifies local high-density regions.
Quantum clustering methods use either the paradigm of AQC~\cite{bauckhage_ising_2018,arthur_balanced_2021, nguyen_quantum_2023,kumar_quantum_2018} or circuit-based quantum computing such as Quantum clustering~\citep{NIPS2001_3e9e39fe} that models clustering through the Schr\"odinger equation, where cluster centers are defined as the minima of its potential function. \citet{casana-eslava_probabilistic_2020} extend this formulation with a probabilistic estimate of cluster memberships. These approaches use classical computation and circuit-based quantum computers that are fundamentally different from AQC and currently still on a smaller scale.

Closest to this work, \citet{arthur_balanced_2021} present a balanced k-means clustering algorithm with predefined target size $s_k$ suitable for an AQC and \citet{nguyen_quantum_2023} use a similar formulation to cluster visual features. While our underlying algorithm follows a similar approach as~\cite{arthur_balanced_2021, nguyen_quantum_2023}, they discard all but the best measurement. Our approach, however, utilizes all information by employing AQC as a sampler to generate probabilistic solutions of the clustering problem, rather than only using the best solution in an optimization framework. We particularly consider balanced clustering, as it forms the basis of several AQC algorithms~\citep{benkner_adiabatic_2020,birdal_quantum_2021,zaech_adiabatic_2022} in computer vision.

\paragraph{Uncertainty quantification in clustering.}
Knowing the set of high-quality solutions and their confidence offers valuable insights for both low- and high-level tasks. E.g. at a low level, the probability of top solutions can help to determine the correct cluster count. Choosing a different number than the actual data process introduces ambiguity~\citep{casana-eslava_probabilistic_2019}: too many clusters cause overlap, while too few clusters spread points, reducing associated probability.

For high-level tasks, calibrated clustering solutions have the potential to enhance matching problems like multi-object tracking~\citep{chiu2021probabilistic}. Following the AQC framework, tracking becomes a clustering problem with additional constraints that represent the temporal relation between points~\citep{zaech_adiabatic_2022}, where each cluster corresponds to one track. The feature defining clusters can be visual similarities from re-identification~\citep{hirzer_person_2011,paisitkriangkrai_learning_2015,zheng_joint_2019} or spatial similarity~\citep{bewley_simple_2016}. The set of solutions models different trajectories due to occlusions or intersecting paths. In complex systems like autonomous vehicles, understanding multiple probable solutions helps to predict multimodal future trajectories~\citep{jiang_motiondiffuser_2023, deo_multimodal_2018}, to assess risks when taking actions. In a similar approach, AQC is used for feature matching~\citep{birdal_quantum_2021} and uncertainty estimation potentially allows to discard ambiguous candidates during 3d reconstruction.
\section{Adiabatic Quantum Computing}
Quantum computing is a fundamentally new approach, that utilizes the state of a quantum system to perform computations. In contrast to a classical computer, the state is probabilistic and described by its wave function, which enables the use fundamental properties of quantum systems like superposition and entanglement. Quantum algorithms that can efficiently run on these systems allow to solve previously infeasible problems, including the well-known prime factorization by Shor's algorithm~\citep{shor_polynomialtime_1999}. In the following, the most important basics of quantum computing are explained.

\paragraph{Qubit.}
Qubits are the building block of a quantum computer. In contrast to the classical bit, which is in either of its two basis states $\ket{0}, \ket{1}$, the qubit can exist in a state of superposition, which is a linear combination of any two basis states $\ket{\psi} = \alpha\ket{0} + \beta\ket{1}$.
The two complex-valued factors $\alpha, \beta$ describing the superposition are called amplitudes.

\paragraph{Entanglement.}
To perform meaningful operations, a quantum computer operates on a system of multiple qubits. If the qubits are independent of each other, the joint state of the system can be computed as the tensor-product of all involved qubit states.  If the qubits of a system are entangled~\citep{einstein_can_1935,schrodinger_discussion_1935}, it is is not possible to decompose the joint state into the tensor-product of each separate qubit state. Therefore, measuring the state of one qubit in a entangled system, affects the state of the other qubits. This is a fundamental property of quantum mechanics and plays a crucial role for the capabilities of quantum computing.

\paragraph{Measurement.} During computation on the quantum computer, the state of the system can be any valid superposition of basis states. Nevertheless, a measurement of the system always results in a single state of the measurement basis, which is referred to as wave-function collapse~\citep{wilde_quantum_2017}. The probability of measuring a state is the respective squared amplitude. In the single qubit case, this corresponds to
\begin{equation}
    p(\ket{0}) = |\alpha|^2 \quad\quad p(\ket{1}) = |\beta|^2.
\end{equation}
In contrast to all other operations in quantum computing, this is not reversible.

\paragraph{Adiabatic Quantum Computing.} AQC, which is used in this work, is a quantum computing paradigm where the state of a quantum system is modified by performing an adiabatic transition. Current hardware implementations such as the D-wave systems~\citep{bunyk_architectural_2014} follow this approach by implementing QA to solve quadratic unconstrained binary optimization (QUBO). They are based on the Ising model~\citep{ising_beitrag_1925, kadowaki_quantum_1998}, which describes the configuration of a set of interacting particles that all carry an atomic spin $s_i$. 

The spin can either be +1 or -1 and the particles are coupled by interactions $J_{ij}$ as well as influenced individually by a transversal magnetic field $h_i$. The energy of this system is described by its Hamiltonian function
\begin{equation}
    H(\vct s) = -\sum_{i}\sum_{j} J_{ij}s_is_j - \sum_i h_is_i.
    \label{eq:basics_ising_hamiltonian}
\end{equation}
Thus, finding the lowest energy state or ground state of the Ising model corresponds to solving the QUBO defined by the Hamiltonian function. This relation is used in the QA, where the system of qubits implements an Ising model $H_T$ that represents the QUBO of interest.

The lowest energy state is found by following the adiabatic theorem~\citep{born_beweis_1928}. Starting with an initial system in its ground state described by a Hamiltonian $H_0$, the adiabatic theorem states that during a sufficiently slow change of the Hamiltonian the system never leaves its ground state. The change of the Hamiltonian is called an adiabatic transition and is often performed with a linear schedule
\begin{equation}
    H(t) = H_0 (1-\frac{t}{T_a}) + H_T\frac{t}{T_a}
\end{equation}
over the annealing-time $T_a$ and allows to solve an optimization problem with the Hamiltonian $H_T$.

While in an ideal noise-free case, the system stays in its ground state, any real system is embedded in a temperature bath that can induce a change to a higher energy state. The distribution of measured final states will then follow the Boltzmann distribution with temperature $T$
\begin{equation}
    p(\vct s) = {\mbox{exp}[\nicefrac{-H(\vct s)}{T}]}/{\sum_{\vct s'}\mbox{exp}[\nicefrac{-H(\vct s')}{T}]},
    \label{eq:boltzmann}
\end{equation}
where $p(\vct s)$ describes the probability of finding the system in state $\vct s$ and $\vct s'$ are all possible states. In this work, we use this property, to sample from the Boltzmann distribution corresponding to the energy-based model of clustering.
\section{Balanced Quantum K-Means}
\label{sec:balanced_qcluster}

In the following, we define an \emph{energy-based model} (EBM) that directly connects to AQC. Based on this, we derive the energy function for k-means and demonstrate how an AQC can sample from the corresponding distribution.

\paragraph{EBM Inference.}
We use the variable $Z \in \XSet$, to represent our binary parameters. The posterior density of interest follows from the Bolzmann distribution as:
\begin{align}
\pi(Z) \vcentcolon= p(Z| X) \propto \exp(-E(Z, X)),
\end{align}
where $X$ denotes the observations, a.k.a. the collection of all specified data and $E$ is called the \emph{potential energy}:
\begin{align}
    E(Z, X) \vcentcolon= -(\log p (X|Z) + \log p(Z)).
\end{align}
Following~\cite{birdal2019probabilistic,birdal_quantum_2021}, we consider the probabilistic inference, where we will be interested in the following quantities:\\
\textbf{\RNum 1} Maximum a-posteriori (MAP):
\begin{align}\label{eq:map}
\hat{Z} = \argmax\limits_{Z \in \XSet} \log p(Z | X )
\end{align}
where $\hat{Z}$ denotes the entirety of the sought parameters.\\
\textbf{\RNum 2} The full posterior distribution: $p(Z|X) \propto p(X,Z)$.\\
Both of these problems are very challenging and cannot be directly addressed by standard methods such as gradient descent (problem \RNum 1) or standard MCMC methods (problem \RNum 2). The difficulty in these problems is mainly originated by the fact that the posterior density is non-log-concave (i.e.\ the negative log-posterior is non-convex) and any algorithm that aims at solving one of these problems should be able to operate in the particular space of binary variables. 
Nevertheless, The MAP estimate is often easier to obtain via a Quantum annealer and useful in practice~\cite{birdal_quantum_2021,benkner_adiabatic_2020}. On the other hand, samples from the full posterior can provide important additional information, such as \emph{uncertainty}. Not surprisingly, the latter is a much harder task, especially considering the discrete nature of our problem. Unfortunately, available QAs such as D-Wave~\cite{mcgeoch_dwave_2022} are not directly capable of tackling the second problem since they sample from a modified posterior and not the true one. Our work fills this gap for the task of balanced K-means clustering using quantum annealing as a direct way to sample from a physical system that follows the Boltzmann distribution~\citep{pochart_challenges_2021} and recomputes the posterior from the found solutions to ensure callibration.

\paragraph{Clustering as QUBO.}
We now focus on obtaining a single point-estimate as a solution to problem \RNum 1. 
To solve the clustering problem using AQC, a QUBO formulation of the K-means energy is required. We use a variation of the one-hot encoding approach~\citep{date_qubo_2021} to formulate the QUBO of k-means, with cost terms similar to~\citet{arthur_balanced_2021}. It uses a matrix ${Z} \in \{0,1\}^{K \times I}$ to encode the cluster assignment of $I$ samples to $K$ clusters. Each row corresponds to one of $K$ clusters and each column to one of $I$ samples. An entry ${Z}_{ki} = 1$ indicates that the sample $x_i$ belongs to cluster $c_k$.
As each sample needs to be assigned to a single cluster, the sum of each column of $Z$ needs to satisfy the constraint $\sum_k {Z}_{ki} = 1~\forall i$. The implementation of constrained clustering, where each cluster has a fixed size $s_k$ furthermore requires the row constraints $\sum_i {Z}_{ki} = s_k~\forall k$ on $ Z$.
The total energy of a solution $E(Z,X)$ can be separated into terms $e(i, j, k)$ modeling the energy of assigning the pair of samples $x_i$ and $x_j$ to the same cluster $c_k$.
\begin{equation}
    \begin{aligned}
        &E(Z,X) = \sum_k \sum_i \sum_j Z_{ki}  Z_{kj} e(i, j, k) + E(Z)
        \label{eq:opt_quadratic_contrained}
    \end{aligned}
\end{equation}
where $E(Z)$ models the row and column constraints as an indicator function.
By vectorizing $Z$ in row-major order as $\vct z = \text{vec}(Z)$, the energy can be rewritten in matrix form as
\begin{equation}
    E(Z,X) = E(X|Z) + E(Z) = \vct z^\intercal\,\vct{Q}\,\vct z + E(Z),
    \label{eq:opt_quadratic_contrained_matrix}
\end{equation}
where $\vct{Q}$ is a block diagonal matrix with blocks $\vct Q_0, ..., \vct Q_K$ and $\vct{G} \vct{z} = \vct{d}$ corresponds to the matrix formulation of the constraints. Each block $\vct Q_k$ of $\vct{Q}$ is a square matrix that contains the energy $e(i, j, k)$ at $\vct Q_{k,ij}$.

To be solved on the AQC, the potential energy needs to be formulated in a QUBO, which cannot exactly implement $E(Z)$ to model the constraints and is circumvented by using Lagrangian multipliers, leading to the MAP estimate
\begin{equation}
    \hat{\vct{z}} = \argmin_\vct{z} \vct{z}^\intercal\vct{Q'}\vct{z} + \vct{b'}^\intercal\vct{z}
    \label{eq:quadratic_penalty}
\end{equation}
with constraints $\vct{Q'} = \vct{Q} + \lambda\vct{G}^\intercal\vct{G}$ and $\vct{b'} = -2\lambda\vct{G}^\intercal\vct{b}$.
To avoid a mixed discrete and continuous optimization problem, a quadratic penalty reformulation $\lambda||\vct{G}\vct{z} - \vct{d}||_2^2$ is chosen. In contrast to a linear penalty approach, where the multiplier $\lambda$ needs to be optimized, our selection only requires a sufficiently high $\lambda$, as all constraint violations result in an increased penalty term. With such selection, the penalty term evaluates to zero if all constraints are fulfilled, and thus, the minimizer $\hat{\vct{z}}$ of the modified optimization problem is a minimizer of the original optimization problem.
Besides modeling balanced clustering, further constraints can be introduced using Lagrangian multipliers. This allows to represent a wide range of tasks as QUBO clustering problems solvable with AQC in the same way.

\subsection{Probabilistic Quantum Clustering}
The MAP estimate was obtained as the lowest energy solution measured during sampling. 
We now consider approximating the posterior distribution to quantify the confidence of these solutions matching the actual ground truth.
Instead of relying on a sequential Markov Chain Monte Carlo (MCMC) method as done in a plethora of classical algorithms~\cite{del2006sequential}, we will approximate the posterior distribution by recomputing the Boltzmann distribution from the set of likely AQC solutions. As the likely solutions are already sampled during quantum annealing, given that the objective follows the EBM, the posterior distribution can be estimated without any large additional computational overhead. Note, this is unlike MCMC, which incurs significant computational load.

Our clustering formulation employs a mixture of Gaussians to explain the observations and to define the energy function $E(X|Z)$. Each sample in a cluster $c_k$ is modeled as a sample drawn from a Gaussian distribution $\mathcal{N}(\vct{\mu}_k, \vct I)$ with mean $\vct{\mu}_k$ and identity covariance similar to k-means. As shown in Section~\ref{sec:balanced_qcluster}, we are interested in the posterior distribution over possible assignments $Z$:
\begin{equation}
    p(Z | {X}) =
    \frac{p({X}|{Z})p({Z})}{\sum_{{Z}'} p({X}|{Z'})}=
    \frac{p({X}|{Z})p({Z})}{A}.
    \label{eq:method_posterior_abstract}
\end{equation}
While this approach provides a probabilistic estimate by jointly modeling information about the possible cluster configurations and the distribution of data points, it is often intractable to evaluate due to the partition function ${\sum_{{Z}'} p({X}|{Z'})}$, the sum over all possible solutions $Z'$, which incurs exponential cost in the number of samples. To overcome this, we utilize an AQC that samples directly from the corresponding Boltzmann distribution, which we parameterize according to the probabilistic clustering problem.

\paragraph{Data Model.}
Determining the potential energy and thus, cost function of the QUBO is a design choice of the algorithm. For our probabilistic approach, a well-defined data distribution, that forms the basis of the cost function is required. While many tasks approached in quantum computer vision, such as tracking~\citep{zaech_adiabatic_2022} or synchronization~\citep{benkner_qmatch_2021, birdal_quantum_2021}, costs are based on learned metrics or heuristics, they can also be trained to reflect the properties required in our approach.

We therefore, follow the mixture of Gaussian model, where each cluster generates samples from a normal distribution. With the independence of observations and clusters, given the distribution parameters, the likelihood of the joint observations for an assignment $Z$ is given as the product of the individual likelihoods
\begin{align}
    f({X}|{Z}) =
    \prod_{k=1}^{K} f({X}|{Z_k}) =
    \prod_{k=1}^{K} \prod_{i\in Z_k} f({x}_i|{Z_k})
    \label{eq:likelihood_raw}
\end{align}
where the likelihood $f({x}_i|{Z_k})$ corresponds to a Gaussian distribution that follows $\mathcal{N}(\vct{\mu}_k, \vct I)$ and $d$ is the dimensionality of the space. 
This result can be used to formulate the energy-based model with the energy function 
\begin{equation}
    E(X | Z) = \sum_{k=1}^{K} \sum_{i\in Z_k}  \frac{1}{2} ({\vct x_i} - {\vct\mu_k})^\intercal \vct I({\vct x_i} - {\vct\mu_k}).
\end{equation}
Using the energy function to rewrite the posterior distribution leads to the Boltzmann distribution from Equation~\ref{eq:boltzmann}
\begin{equation}
     p(Z | {X}) = \frac{\mbox{exp} \left[ -(E({X} | Z) + E(Z)) \right]}{\sum_{{Z}'} \mbox{exp} \left[ - E({X} | Z') \right]}.
     \label{eq:boltzmann_cluster}
\end{equation}
In this formulation, the energy of the assignment $E(Z)$ corresponds to the prior $p({Z})$, which models feasible and infeasible solutions by an indicator function. In the case of balanced clustering, this corresponds to allowing assignments that have each point assigned to exactly one cluster and a cluster size according to $s_k$.
When the AQC is used as an optimizer, finding the the lowest energy solution corresponds to the MAP as shown in Equation~\ref{eq:map}

As the AQC qubit system is an Ising model, it requires formulating $E(X | Z)$ and $E(Z)$ quadratic in the optimization variables $Z$, enabling the joint discovery of assignments and cluster means. This is achieved by using the maximum likelihood estimator of the mean $\mu_k=\frac{1}{s_k} \sum_i Z_{ki} x_i$, resulting in a quadratic energy formulation that fits the Ising model in Equation~\ref{eq:quadratic_penalty}
\begin{equation}
    E_k({X} | Z) 
    = \frac{1}{s_k} \sum_i \sum_j Z_{ki} Z_{kj}~(\vct x_i - \vct x_j)^\intercal (\vct{x}_i - \vct x_j).
    \label{eq:quadratic_energy}
\end{equation}

The second energy term  $E(Z)$ in Equation~\ref{eq:boltzmann_cluster}, modeling the prior distribution over possible assignments cannot exactly be embedded in the Ising model and is approximated using Lagrange multipliers. Importantly, this does not influence the energy of feasible solutions relevant in the posterior distribution, as the penalty evaluates to zero for these.

\paragraph{Posterior recomputation.}
While ideally the measurements on AQC are direct samples from the Boltzmann distribution~\citep{d-wavesystemsinc._performance_2017,dixit_training_2021}, it requires solving a range of challenges that prohibit a direct use in our scenario~\citep{pochart_challenges_2021}: Mapping between the cost function and the physical system implemented on the AQC requires Hamiltonian scaling~\citep{pochart_challenges_2021}. Estimating the scaling factor is nontrivial~\citep{nelson_highquality_2022} and prohibitive for using samples directly in many cases. Additionally, hardware limitations including imperfection of the processor and the spin-bath polarization effect~\citep{pochart_challenges_2021} prevent the AQC to sample the Boltzmann distribution exactly. Finally, as the energy term $E(Z)$ can only be implemented using the penalty method, the sampling density is influenced by solutions not fulfilling the constraints.
We compensate for these limitations by evaluating the energy of all measured feasible solutions $Z'$ and recompute $p(Z | {X})$ by evaluating the analytic partitioning function over these, which only requires sampling solutions at a sufficiently high temperature.

\paragraph{Coresets.}
Having the set of the most likely clustering solutions $Z$ available together with their posterior probability $P(Z | X)$ allows to find an assignment $Z^*$ of a subset of points that solves the clustering problem with increased probability $P(Z^* | X)$. Such a set can be found by using Algorithm~\ref{alg:maxset_search}, which implements a greedy approach that disregards points that disagree between different clustering solutions.
With a sufficiently well-sampled Boltzmann distribution, the resulting coreset assignment $Z^*$ with the corresponding probabilistic estimate $P(Z^* | X)$ is still calibrated.
\begin{algorithm}[t]
\begin{algorithmic}[1]
\State $Z^* \gets Z_0$
\State $p \gets P(Z_0 | \vct X)$
\State $i \gets 1$
\While{$p \leq p_\text{min}$}
\State $Z_i' \gets  align(Z_i, Z^*)$
\LeftComment{Find the cluster permutation that minimizes the number of points assigned to different clusters in $Z_i$ and $Z^*$.}
\State $Z^* \gets Z^* \cap Z_i'$
\LeftComment{Remove points assigned to different clusters.}
\State $p \gets p + P(Z_i |  X)$
\State $i \gets i + 1$
\EndWhile
\State \textbf{return} $Z^*$
\end{algorithmic}
\caption{CoresetSearch}
\label{alg:maxset_search}
\end{algorithm}

\paragraph{Lagrange multiplier optimization.}
Using the quadratic penalty method to implement the constraints requires finding suitable Lagrangian multipliers~$\lambda$. Even though a very high multiplier theoretically guarantees finding a feasible solution, it also deteriorates the conditioning of the optimization problem. Therefore, a suitable Lagrangian multiplier lifts the cost of any constraint violation above all relevant solutions of the clustering problem, while keeping them low enough to avoid scaling the total energy of the problem up considerably. To estimate the multipliers, we follow an iterative procedure as also proposed in~\citep{zaech_adiabatic_2022}.

In an initial step, balanced classical k-means~\citep{malinen_balanced_2014} is used to find a feasible clustering solution. This solution is used to estimate Lagrangian multipliers that avoid any first-order violation of the constraints. In subsequent iterative optimization steps, the problem is solved in simulation, to find multipliers that result in a well-conditioned problem.

Such optimization procedure is crucial due to the low fidelity of the current generation of QAs, which requires careful engineering of the problem energy. Therefore, we expect this procedure to become of reduced importance with the progress of quantum computing.
\section{Experiments and Results}
We perform experiments on synthetic as well as real data to verify the efficacy of our method in finding the set of high-probability solutions and in estimating calibrated confidence scores. The experimental scenarios are solved with QA, Simulated Annealing (SIM), and exact exhaustive search using the presented energy formulation and with k-means as a baseline method.
This further allows us to understand the limitations and required work when deploying the approach to real quantum computers.

\begin{figure}[t!]
    \centering\vspace{2mm}
    \begin{subfigure}[b]{0.30\columnwidth}\vspace{-5mm}
        \raisebox{4mm}{\includegraphics[width=\linewidth, trim={5 0 24 0}, clip]{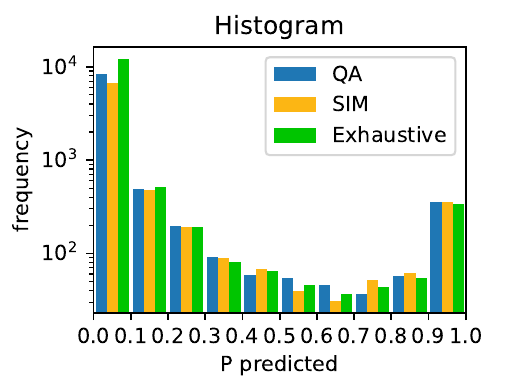}}
    \end{subfigure}
    \hfill
    \begin{subfigure}[b]{0.68\columnwidth}
        \includegraphics[width=\linewidth, clip]{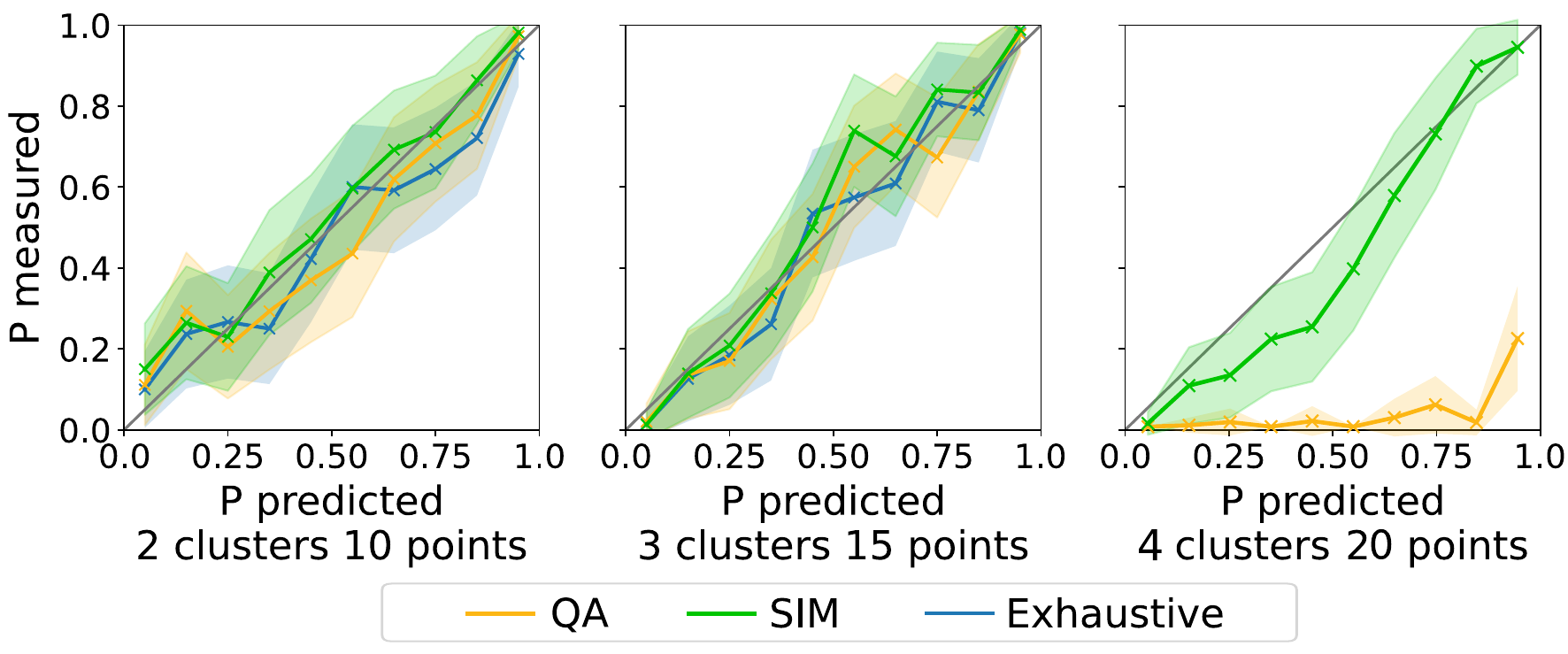}
    \end{subfigure}\vspace{-3mm}
    \caption{Evaluation of the calibration and distribution for QA, SIM and exhaustive search in tasks with 2 to 4 clusters, each with 5 points. All results are generated with 1,000 problems in each scenario and 5,000 measurements for each clustering problem.\vspace{-3mm}}
    \label{fig:hist_cal}
\end{figure}

\paragraph{Implementation details.}
\noindent\textbf{Quantum annealing} (QA) experiments are performed on the D-Wave Advantage 2 Prototype 1.1~\citep{mcgeoch_dwave_2022}. The system offers 563 working qubits, each connected with up to 20 neighbors. For each clustering problem 5000 measurements are performed, each with 50$\mu\text{s}$ annealing time. 
Due to the strong compute-time limits on an QA, all Lagrangian optimization steps are performed with SIM, before measuring the final results on the QA.

\noindent\textbf{Simulated annealing} (SIM) provided by D-Wave is used for larger scale comparisons. Similar to QA, we perform 5000 runs for each clustering problem. We reduce the number of sweeps performed in each run to 30, which allows us to sample the Boltzmann distribution at a sufficiently high temperature in most scenarios, making it comparable to QA.

\noindent\textbf{Exact exhaustive search} is used as a reference method to validate the energy-based formulation on small problems. By iterating all feasible solutions, the lowest energy solution is guaranteed to be found and the partitioning function is computed exactly.

\noindent\textbf{K-Means clustering} with a balanced cluster constraint~\citep{malinen_balanced_2014} forms the baseline for our approach. We run the algorithm until convergence for a maximum of 1000 iterations. While this solution does not provide a probabilistic estimate, it is useful to assess the relative clustering performance.

\newcommand{\rotationangle}{90}
\newcommand{\tablemetrics}{
\rotatebox{\rotationangle}{\parbox{2.2cm}{\centering Accuracy}} & 
\rotatebox{\rotationangle}{Completeness} & 
\rotatebox{\rotationangle}{\parbox{2.2cm}{\centering Adjusted\\Rand~Index}} &
\rotatebox{\rotationangle}{\parbox{2.2cm}{\centering Fowlkes-\\Mallows Score}}}
\begin{table*}[tb]
    \centering
    \resizebox{1.0\linewidth}{!}{
    \begin{tabular}{l cccc@{\hskip 8mm}cccc@{\hskip 8mm}cccc}
        \rotatebox{\rotationangle}{\parbox{2.2cm}{\centering Solver\\Method}}
        & \tablemetrics & \tablemetrics & \tablemetrics  \\
        
        \midrule
        & \multicolumn{4}{c}{15 Points, 3 Clusters, 2 Dim} & 
        \multicolumn{4}{c}{\hspace{-0mm}30 Points, 3 Clusters, 2 Dim} & 
        \multicolumn{4}{c}{\hspace{-0mm}45 Points, 3 Clusters, 2 Dim} \\
        
        SIM 
        & \textbf{56.4}$\pm$1.6 & \textbf{79.5}$\pm$0.8 & \textbf{74.7}$\pm$1.0 & \textbf{81.9}$\pm$0.7 
        & \textbf{38.6}$\pm$1.5 & \textbf{74.2}$\pm$0.8 & \textbf{73.3}$\pm$0.9 & \textbf{81.6}$\pm$0.6 
        & {50.8}$\pm$1.6 & \textbf{86.2}$\pm$0.5 & \textbf{87.3}$\pm$0.5 & \textbf{91.4}$\pm$0.3 \\ 
        K-means 
        & 51.3$\pm$1.6 & 75.0$\pm$0.9 & 68.8$\pm$1.1 & 77.7$\pm$0.8 
        & {37.0}$\pm$1.5 & {71.9}$\pm$0.8 & {70.2}$\pm$0.9 & {79.4}$\pm$0.6 
        & \textbf{52.8}$\pm$1.1 & {85.9}$\pm$0.4 & {86.6}$\pm$0.4 & {90.9}$\pm$0.3 \\ 
        
        \midrule
        & \multicolumn{4}{c}{15 Points, 3 Clusters, 2 Dim}
        & \multicolumn{4}{c}{\hspace{-0mm}10 Points, 2 Clusters, 2 Dim}
        & \multicolumn{4}{c}{\hspace{-0mm}20 Points, 4 Clusters, 4 Dim} \\
        
        QA 
        & 56.1$\pm$1.6 & 79.4$\pm$0.8 & 74.6$\pm$1.0 & \textbf{81.9}$\pm$0.7 
        & \textbf{74.3}$\pm$1.4 & \textbf{80.2}$\pm$1.1 & \textbf{79.6}$\pm$1.1 & \textbf{88.7}$\pm$0.6 
        & {12.4}$\pm$1.0 & {51.2}$\pm$0.8 & {34.0}$\pm$1.0 & {47.9}$\pm$0.8 \\ 
        SIM 
        & \textbf{56.4}$\pm$1.6 & \textbf{79.5}$\pm$0.8 & \textbf{74.7}$\pm$1.0 & \textbf{81.9}$\pm$0.7 
        & 74.1$\pm$1.4 & 80.0$\pm$1.1 & 79.5$\pm$1.1 & 88.6$\pm$0.6 
        & \textbf{32.4}$\pm$1.5 & \textbf{70.4}$\pm$0.8 & \textbf{59.2}$\pm$1.1 & \textbf{67.8}$\pm$0.8 \\ 
        K-means 
        & 51.3$\pm$1.6 & 75.0$\pm$0.9 & 68.8$\pm$1.1 & 77.7$\pm$0.8 
        & {70.1}$\pm$1.4 & {76.3}$\pm$1.2 & {75.4}$\pm$1.2 & {86.3}$\pm$0.7 
        & {20.9}$\pm$1.3 & {61.9}$\pm$0.8 & {47.4}$\pm$1.0 & {58.5}$\pm$0.8 \\ 
        Exhaustive 
        & \textbf{56.4}$\pm$1.6 & \textbf{79.5}$\pm$0.8 & \textbf{74.7}$\pm$1.0 & \textbf{81.9}$\pm$0.7 
        & \textbf{74.3}$\pm$1.4 & \textbf{80.2}$\pm$1.1 & \textbf{79.6}$\pm$1.1 & \textbf{88.7}$\pm$0.6 
        & - & - & - & - \\ 
        
        \bottomrule
    \end{tabular}}
    \caption{Synthetic data results for our approach (QA, SIM, exhaustive) and k-means. All numbers in \% with standard error of the mean.\vspace{-3mm}}%
    \label{table:comparison_clustering_metrics}\vspace{-3mm}
\end{table*}

\noindent\textbf{Data} for the quantitative evaluation of our method is synthetically generated. For each clustering problem a total of $I$ points are sampled from a separate normal distribution for each of $K$ clusters. The centroids are randomly drawn, such that the distance between each pair of clusters lies within a predefined range $[d_\text{min}, d_\text{max}]$. For each experiment a total of $L$ clustering tasks is generated. This allows us to evaluate the calibration metrics over a large value range. For all experiments that directly compare methods, identical clustering tasks are used.

Further results are provided for the IRIS dataset~\citep{fisher_use_1936} and a dataset of images containing ambiguous objects that are to be identified. The IRIS contains 50 samples of 4 features in 3 classes. We randomly subsample the points and dimensions to generate the parameters required for our experiments.

\noindent\textbf{Clustering metrics} are computed using the available ground-truth clusters. We evaluate 4 standard metrics: The accuracy, which measures the ratio of clustering solutions that are identical to the ground-truth. Completeness~\citep{rosenberg_vmeasure_2007} measures the ratio of points from a single cluster being grouped together. The adjusted Rand score~\citep{hubert_comparing_1985} compares all pairs of points in the ground truth and prediction and the Fowlkes-Mallows index~\citep{fowlkes_method_1983} combines precision and recall into a single score.

\subsection{Results}
\paragraph{Calibration performance.}
We evaluate the calibration of our method on a synthetic clustering scenario with 3 clusters and 15 points using QA, SIM and exhaustive search on 1000 tasks. First all clustering solutions $Z$ are accumulated in bins according to their estimated posterior probability $P(Z|X)$. This process also includes all sampled non-optimal but feasible solutions. Figure~\ref{fig:hist_cal} shows the resulting histogram of solutions and the ratio of correct solutions in each bin after accumulation. It thus evaluates calibration, where the ideal case has mean predicted probabilities on the diagonal. We find our approach to generate well-calibrated probabilities in both simulation and when using QA for the two smaller experiments for up to 15 points. Due to the problem QUBO size, only SIM is able to find the correct solutions on the largest example.

\paragraph{Clustering performance.}
We evaluate the performance of our clustering formulation using synthetic data in Table~\ref{table:comparison_clustering_metrics}. The upper rows show results for 3 clusters with an increasing number of points and 1000 tasks/setting for SIM and k-means. Due to the problem size, it cannot be solved using QA and exhaustive search. Our formulation with SIM outperforms k-means on the small tasks, however, the difference vanishes with increasing problem size. This can be attributed to the globally optimal solution our approach is optimizing for. Ideally, the global solution is at least as good as the k-means solution, however, an increasing problem size also increases the complexity of the QUBO, which explains the dropping performance on large problems.

For the largest clustering problem with 3 clusters and  45 points, k-means provides the best accuracy with more correct solutions compared to SIM. However, the clustering quality metrics are higher for SIM. This indicates that our formulation is able to find a better solution in cases where the problem is not solved correctly by SIM and k-means.

\begin{figure}[tb]
  \centering
  \subfloat[\centering Solver comparison for the adjusted Rand index.
    \label{fig:sparsification_method_comp}]{\includegraphics[height=35mm, trim={15 0 0 33},clip]{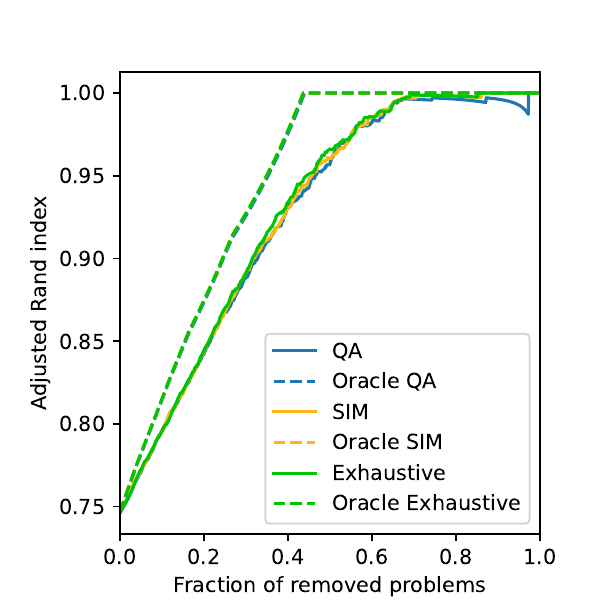}}
    \subfloat[\centering Metric comparison for the QA solver.\label{fig:sparsification_metric_comp}]
    {\includegraphics[height=35mm, trim={-5 0 35 33},clip]{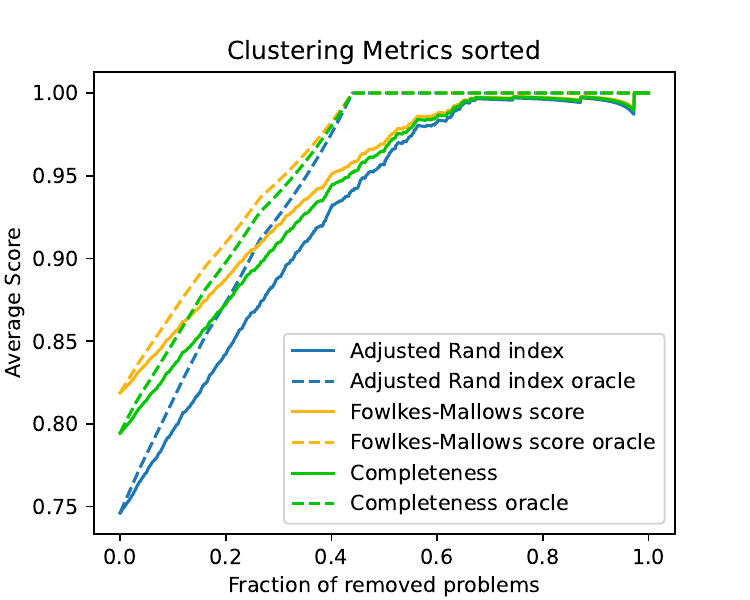}}
  \caption{Sparsification plots of clustering metrics.\vspace{-3mm}}
\end{figure}

The lower rows in Table~\ref{table:comparison_clustering_metrics} show settings with 2,3, and 4 clusters, each containing 5 points, again with 1000 tasks/setting. For the two smaller scenarios QA on the D-wave Advantage 2 provides results close or identical to SIM and exhaustive search. For the largest scenario with 20 points in 4 clusters, QA loses performance.

The quality of predicted posterior probabilities can be further evaluated by linking them to the clustering metrics and sparsifying the set of tasks. Starting with metrics over the whole set of clustering tasks, we remove tasks according to increasing probability and evaluate the metric over the remaining set. For an ideal predictor, this removes the lowest-performing tasks first. In Figure~\ref{fig:sparsification_method_comp} this is done for the adjusted Rand index with \textit{different solvers} and in Figure~\ref{fig:sparsification_metric_comp} QA with \textit{different metrics}. The solid lines show the sparsification plots using the predicted probabilities and the dashed lines show the same plots for an oracle method that generates the best possible ordering based on the metrics.

In Figure~\ref{fig:sparsification_method_comp} it becomes apparent that QA, SIM and exhaustive search perform close to each other over most of the value range. Nevertheless, for a high sparsification with more than 80\% of the tasks removed, QA shows a drop in performance compared to the other methods. This is caused by tasks where only a single, but incorrect solution is found, which gets assigned a posterior probability of $P(Z|X) = 1.0$. In such cases, the Boltzmann distribution has not been sampled sufficiently well, either because of too few measurements or because of a low effective sampling temperature. As SIM does not show this behavior the source can likely be traced back to current limitation of the quantum computer.

\begin{figure}
    \captionsetup[subfigure]{labelformat=empty}
    \begin{subfigure}{0.25\linewidth}
        \centering
        \includegraphics[width=0.8\linewidth, trim={20 20 396 22}, clip]{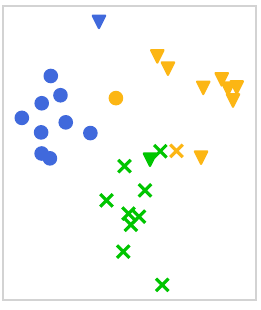}
        \caption{$p=0.61$}
    \end{subfigure}%
    \begin{subfigure}{0.25\linewidth}
        \centering
        \includegraphics[width=0.8\linewidth, trim={146 20 270 22}, clip]{figures/maxset_qual_ICCV_3C_3O_10P_2D_1STD_SIM_wide.pdf}
        \caption{$p=0.77$}
    \end{subfigure}%
    \begin{subfigure}{0.25\linewidth}
        \centering
        \includegraphics[width=0.8\linewidth, trim={273 20 143 22}, clip]{figures/maxset_qual_ICCV_3C_3O_10P_2D_1STD_SIM_wide.pdf}
        \caption{$p=0.86$}
    \end{subfigure}%
    \begin{subfigure}{0.25\linewidth}
        \centering
        \includegraphics[width=0.8\linewidth, trim={400 20 16 22}, clip]{figures/maxset_qual_ICCV_3C_3O_10P_2D_1STD_SIM_wide.pdf}
        \caption{$p=0.90$}
    \end{subfigure}
    \caption{Visualization of coresets for synthetic data with the probability for each of the determined pointsets.\vspace{-2mm}}
    \label{fig:max_pointset_sim}
\end{figure}

\paragraph{Coresets.}
Qualitative examples for the coresets generated with Algorithm~\ref{alg:maxset_search} are depicted in Figure~\ref{fig:max_pointset_sim}. The shape of each point represents the ground truth class and the color the assigned cluster. Starting with the most likely solution having a predicted probability of $p(Z|X) = 0.61$ on the left, each plot shows one additional step of the algorithm, which successively removes points from the solution, indicated by plotting them in Grey. The illustration demonstrates that the CoresetSearch algorithm is able to generate well-separated clusters from the probabilistic predictions.

\begin{figure}[tb]
    \centering
    \begin{floatrow}
        \floatbox[\nocapbeside]{figure}[0.45\textwidth]{%
            \subfloat[QA.
            \label{fig:qual_iris_qa}]{
            \includegraphics[width=0.45\linewidth, trim={26 42 20 48},clip,valign=t]{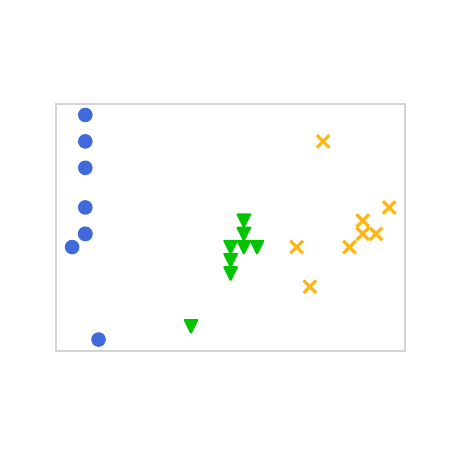}
            \vphantom{\includegraphics[width=0.45\linewidth, trim={26 30 20 35},clip,valign=t]{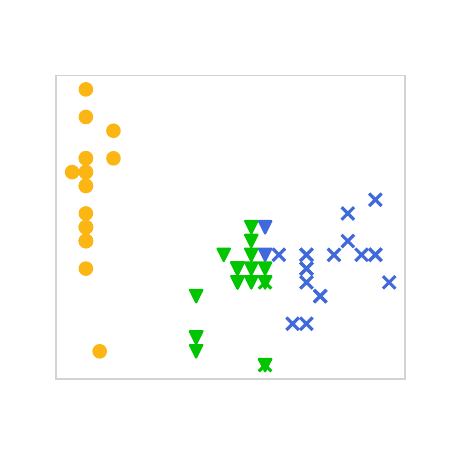}}
            }
            \hfill
            \subfloat[SIM.
            \label{fig:qual_iris_sim}]{
            \includegraphics[width=0.45\linewidth, trim={26 30 20 35},clip,valign=t]{figures/qual_IRIS_SIM.pdf}}
        }{%
            \caption{Qualitative results on the IRIS dataset.}
        }
        
        \floatbox[\nocapbeside]{table}[0.45\textwidth]{%
                \caption{Performance on IRIS subsets (3 clusters/15 points).}
                \label{table:comparison_clustering_metrics_iris}
            }{%
                \centering%
                \resizebox{1.0\linewidth}{!}{
                \begin{tabular}{lcccc}
                    & \tablemetrics \\
                    \midrule
                    QA 
                    & 47.2 & 81.7 & 76.8 & 83.4\\
                    SIM 
                    & 47.1 & 81.7 & 76.7 & 83.4 \\
                    K-means 
                    & 47.0 & 80.6 & 75.5 & 82.5 \\
                    Exhaustive 
                    & 47.1 & 81.8 & 76.8 & 83.5 \\
                    \bottomrule
                \end{tabular}}
            }
    \end{floatrow}\vspace{-3mm}
\end{figure}

\begin{figure}
    \centering\includegraphics[width=\linewidth, trim={28 28 18 25},clip]{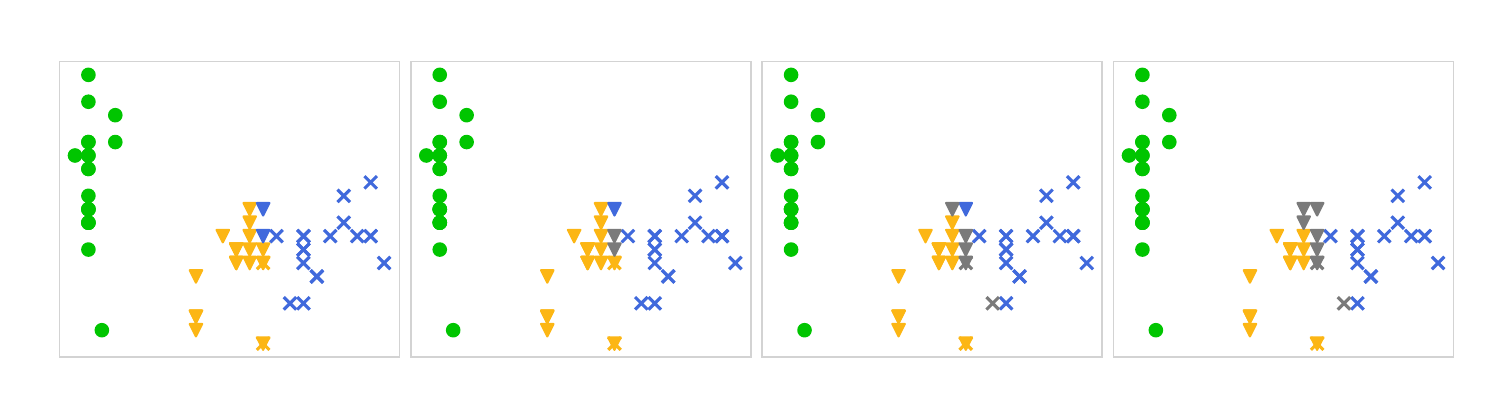}
    \caption{Coresets generated using SIM on the IRIS dataset with 60 points.\vspace{-3mm}}%
    \label{fig:max_pointset_iris}
\end{figure}

\begin{figure}
    \centering\includegraphics[width=\linewidth, trim={0 25 0 35 },clip]{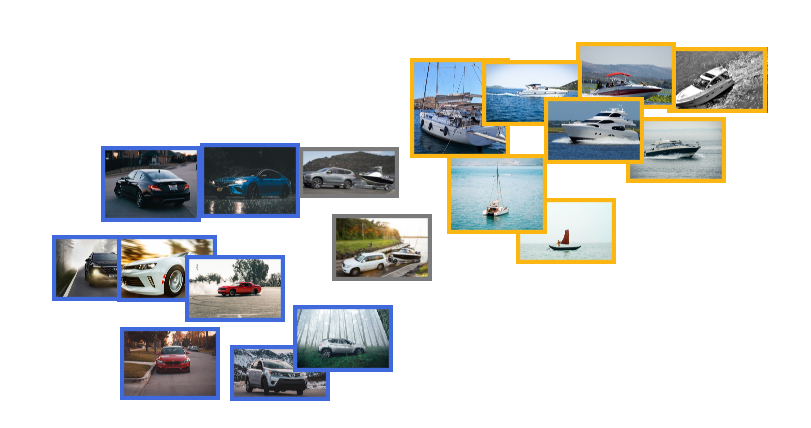}
    \caption{First coreset using SIM on our collected dataset.\vspace{-3mm}}%
    \label{fig:max_pointset_carboat}
\end{figure}

\paragraph{Real-World Datasets}
While our method assumes an identity covariance in the data, it can be applied to other distributions, which we evaluate on the widely used IRIS dataset as well as a set of self-collected images. Clustering metrics for experiments on IRIS using 3 clusters with 5 points each are provided in Table~\ref{table:comparison_clustering_metrics_iris} and show that our formulation using QA and SIM is competitive with k-means. Figures~\ref{fig:qual_iris_qa} and~\ref{fig:qual_iris_sim} show differently sized qualitative examples from the dataset solved using our formulation with QA and SIM respectively. On the full IRIS dataset, SIM and k-means achieve identical results with a Completeness of 77.7\%, Adjusted Rand index of 78.6\% and further results provided in the supplementary. The results from Algorithm~\ref{alg:maxset_search} using results from SIM are provided in Figures~\ref{fig:max_pointset_iris} and \ref{fig:max_pointset_carboat} for IRIS and our dataset respectively. They show that even for a distribution mismatch the coresets can provide meaningful results for removing ambiguous samples.

To demonstrate that meaningful ambiguous samples can be identified in a high-dimensional space present in image data, we collected publicly available images of cars, boats and cars towing boats. Image features were extracted using VGG~\cite{simonyanVeryDeepConvolutional2014} and subsequently clustered using our approach. Similar to IRIS, the distribution does not match our assumptions, however, we are still able to identify ambiguous images that contain cars towing a boat by using Algorithm~\ref{alg:maxset_search}. The first coreset is depicted in Figure~\ref{fig:max_pointset_carboat} where differently colored frames indicate the cluster assignment and Grey frames mark ambiguous samples. Clustering is performed on the 1000-dimensional feature and images are plotted at their 2D projection generated using UMAP~\cite{mcinnes_umap_2018}. Our approach correctly identifies the images that contain a car towing a boat and thus, cannot be uniquely assigned to either cluster.

\section{Conclusion}
In this work, we proposed a probabilistic clustering approach based on sampling k-means solutions using AQC. By using all valid measurements, calibrated confidence scores are computed at little cost and solutions are competitive to an iterative balanced k-means approach. We evaluated our method on synthetic as well as real data using simulation, exhaustive search as well as the D-Wave Advantage 2 prototype QA to explore the potential of quantum computing in machine learning and computer vision.

While the approach is still limited in the problem size, quantum computing enables a fundamentally different approach to clustering that can provide additional information that is costly to compute otherwise. Nevertheless, even with the current progress and potential to scale to real-world problems, more work is required to adapt existing problem formulations, such that the full capability of quantum computing can be effectively used.

\paragraph{Acknowledgement:} This work was funded by Toyota Motor Europe via the research project TRACE Z\"urich.


\end{document}


\maketitle

\section{Introduction}
\label{sec:intro}

The main manuscript presents a novel approach for probabilistic clustering based on exploiting the probabilistic nature of adiabatic quantum computing (AQC). In the supplementary material, we provide additional details and that complement the main manuscript. Section~\ref{sec:energy_derivation} presents the detailed derivation of the energy function used in the manuscript. Section~\ref{sec:inference_parameter_optimization} discusses the optimization of the inference parameters to increase the AQC solver performance. Section~\ref{sec:data_gen} provides information on data generation for the synthetic and real-world datasets used in the experiments. Sections~\ref{sec:clustering_performance}, \ref{sec:coreset_sparsification_performance} and \ref{sec:performance_iris} extend the clustering performance and calibration evaluation on synthetic data and the IRIS dataset respectively. In Section~\ref{sec:failure_cases}, we discuss failure cases encountered during the experiments. Finally, we outline the limitations of our approach in Section~\ref{sec:limitations}.

Overall, the supplementary material aims to clarify open questions from the main manuscript, provides additional insights into the proposed method and discusses its current limitations.

\section{Energy Function Derivation}
\label{sec:energy_derivation}

The following section shows the step-by-step derivation of the energy function used in this paper. As clusters are independent, the energy for each cluster can be computed separately $E({X} | Z) = \sum_k E_k({X} | Z)$, where ${X}$ represents the data-points and $Z$ is the assignment matrix with entry $Z_{ki}$ assigning point $x_i$ to cluster $c_k$. For a single cluster $c_k$, the energy can be further extended into the quadratic and linear terms as follows

\begin{equation}
    \begin{aligned}
        &E_k({X} | Z) = \sum_i Z_{ki}({\vct x_i} - {\vct \mu_k})^\intercal \vct{I} ({\vct x_i} - {\vct \mu_k})\\
        &= \sum_i Z_{ki}(\vct x_i^\intercal \vct{I} \vct x_i
        -2 \vct x_i^\intercal \vct{I} \vct \mu_k
        + \vct \mu_k^\intercal \vct{I} \vct \mu_k)\\
        &= \sum_i Z_{ki}\vct x_i^\intercal \vct{I} \vct x_i
        -2 \sum_i Z_{ki} \vct x_i^\intercal \vct{I} \vct \mu_k
        + \sum_i Z_{ki} \vct \mu_k^\intercal \vct{I} \vct \mu_k\\
        &= \sum_i Z_{ki}\vct x_i^\intercal \vct{I} \vct x_i
        -2 \sum_i Z_{ki} \vct x_i^\intercal \vct{I} \vct \mu_k
        + s_k \vct \mu_k^\intercal \vct{I} \vct \mu_k,
    \end{aligned}
    \label{eq:quadratic_energy1}
\end{equation}
with the cluster mean $\vct{\mu_k}$ and identity matrix $\vct{I}$. By using the maximum likelihood~(ML) estimator of the cluster mean
\begin{equation}
    \vct \mu_k = \frac{1}{s_k} \sum_j Z_{kj} \vct x_j,
\end{equation}
with cluster size $s_k$, the energy formulation only depends on the data and the cluster assignment
\begin{equation}
    \begin{aligned}
        &\sum_i Z_{ki}\vct x_i^\intercal \vct{I} \vct x_i
        -2 \sum_i Z_{ki} \vct x_i^\intercal \vct{I} \vct \mu_k
        + s_k \vct \mu_k^\intercal \vct{I} \vct \mu_k\\
        &= \sum_i Z_{ki}\vct x_i^\intercal \vct{I} \vct x_i
        -2 \sum_i Z_{ki}\vct x_i^\intercal \vct{I} \frac{1}{s_k} \sum_j Z_{kj} \vct x_j\\
        &+ s_k \frac{1}{s_k} \sum_i Z_{ki} \vct x_i^\intercal \vct{I} \frac{1}{s_k} \sum_j Z_{kj} \vct x_j\\
        &= \sum_i Z_{ki}\vct x_i^\intercal \vct{I} \vct x_i
        - \frac{1}{s_k} \sum_i \sum_j Z_{ki} Z_{kj} \vct x_i^\intercal \vct{I} \vct x_j.
    \end{aligned}
    \label{eq:quadratic_energy2}
\end{equation}
This finally shows that the total energy only depends on the distance between each pair of points

\begin{equation}
    \begin{aligned}
        &\sum_i Z_{ki}\vct x_i^\intercal \vct{I} \vct x_i
        - \frac{1}{s_k} \sum_i \sum_j Z_{ki} Z_{kj} \vct x_i^\intercal \vct{I} \vct x_j\\
        &= \frac{1}{s_k} \sum_i \sum_j Z_{ki} Z_{kj}~\vct x_i^\intercal \vct{I} \vct x_i - \vct x_i^\intercal \vct{I} \vct x_j\\
        &= \frac{1}{s_k} \sum_i \sum_j Z_{ki} Z_{kj}~\vct x_i^\intercal \vct{I} (\vct x_i - \vct x_j)\\
        &= \frac{1}{s_k} \sum_i \sum_j Z_{ki} Z_{kj}~\frac{1}{2}[\vct x_i^\intercal \vct{I} (\vct x_i - \vct x_j) + \vct x_j^\intercal \vct{I} (\vct x_j - \vct x_i)]\\
        &= \frac{1}{s_k} \sum_i \sum_j Z_{ki} Z_{kj}~(\vct x_i - \vct x_j)^\intercal \vct{I} (\vct x_i - \vct x_j).\\
    \end{aligned}
    \label{eq:quadratic_energy3}
\end{equation}

\section{Inference Parameter Optimization}
\label{sec:inference_parameter_optimization}

Using the quadratic penalty method to include constraints in the Quadratic Binary Optimization (QUBO) formulation requires finding suitable Lagrangian multipliers $\lambda$. Even though a very high multiplier theoretically guarantees to find a feasible solution, it also deteriorates the conditioning of the optimization problem. Therefore, a suitable Lagrangian multiplier lifts the cost of any constraint violation above all relevant solutions of the clustering problem, while keeping them low enough to avoid scaling the total energy of the problem up considerably. To estimate the multipliers for each constraint, we follow an iterative procedure.

In an initial step, balanced k-means\cite{malinen_balanced_2014} is used to find a feasible clustering solution. This is used to offset the distance terms for each point such that the total clustering solution has an energy of 0. For the next steps, the constraints are separated into 3 components:

The cluster size constraint is defined by $\sum_i {Z}_{ki} = s_k~\forall k$. Due to its strong diagonal term in its quadratic form using Lagrange multipliers, it quickly degrades the energy scaling of the problem. The Lagrangian multiplier corresponding to this constraint is estimated from the maximum cost improvement that can be achieved by switching one point between clusters.

The constraint $\sum_k {Z}_{ki} = 1~\forall i$, which ensures the matching of every point to exactly one cluster, is further segmented into two parts. One part contains the positive off-diagonal elements and penalizes assigning a single point to multiple clusters. Its corresponding Lagrangian is computed from the maximum cost improvement that can be achieved by assigning an additional point to any cluster, compared to the k-means solution. The other term contains negative diagonal elements and adds an incentive to assign every point to one cluster. The corresponding multiplier is estimated by the maximum cost improvement by removing one point from a cluster and thus violating the constraint. 

In five subsequent optimization steps the clustering problem is solved using simulated annealing and the Lagrangian multipliers are increased for constraints that are not fulfilled.

In the last step, which is only performed for simulated annealing, measurements at low temperatures of the Boltzmann distribution are handled. In scenarios where only a single valid clustering solution is returned, the Lagrangian multiplier of the cluster size constraint is increased, which results in sampling the problem at a higher temperature.

Finding well-suited Lagrangian multipliers is crucial due to the low fidelity of the current generation of AQCs, which requires careful engineering of the problem energy. Therefore, we expect this procedure to become of reduced importance in future generations of lower noise AQCs.

For experiments performed on the D-Wave AQC, all Lagrangian optimization steps are performed with SIM, before measuring the final results on the AQC due to the strong compute-time limitations.

\section{Data Generation}
\label{sec:data_gen}

\noindent\textbf{Synthetic Data} is generated by sampling a total of $I$ points from separate normal distributions for each of $K$ clusters. The cluster centers are selected as the corners of a simplex with uniformly drawn edge length, such that the distance between each pair of clusters lies within a predefined range $[d_\text{min}, d_\text{max}]$. The feature-space needs to be at least $K-1$ dimensional for each clustering problem. Sampling the edge-length randomly allows to generate a wide range of clustering problems with a different degree of ambiguity, due to the changing degree of overlap between distributions. This allows to evaluate the whole range of predicted posterior probability values. For each experiment a total of $L$ clustering tasks is generated to evaluate the clustering metrics.

\noindent\textbf{IRIS}~\cite{fisher_use_1936} is subsampled to generate quantitative results over different clustering scenarios. The whole IRIS dataset contains 3 classes, 50 samples for each class and 4 features forming a 4-dimensional space. According to the experiment parameters we randomly select a subset of classes, samples and features without replacement to allow running the tasks on a D-wave AQC. This generates different clustering problems, while keeping the general structure of the data in IRIS.

\noindent\textbf{Image data} is used to demonstrate the applicability of our method to computer vision tasks. We collected 8 images each for cars and boats and two images of cars towing boats. Visual embeddings for each image are extracted after the last layer of VGG16~\cite{simonyanVeryDeepConvolutional2014} pretrained on Imagenet. Subsequently, the high-dimensional features are clustered using our approach, and the first coreset is computed to identify the ambiguous samples as demonstrated in the main paper.

\begin{figure}[tb]
  \centering
  \subfloat[\centering 3 clusters, 30 points.
    \label{fig:calibration_3310}]{
    \includegraphics[width=0.48\linewidth, trim={0 0 25 32},clip]{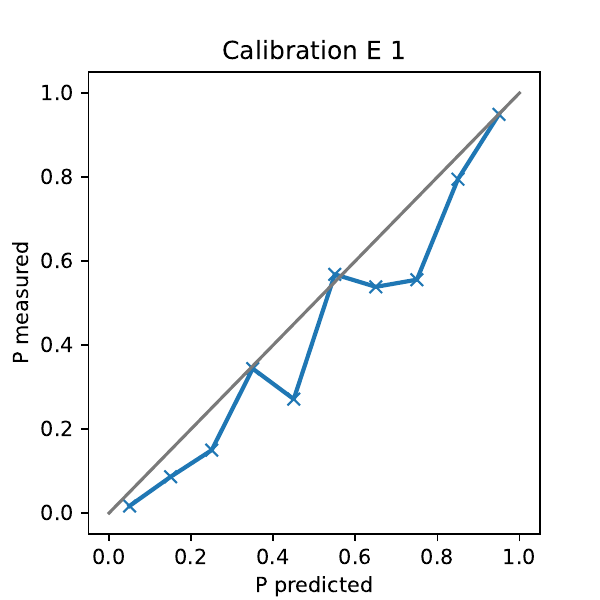}}
  \hfill
  \subfloat[\centering 3 clusters, 45 points.
    \label{fig:calibration_3315}]{
    \includegraphics[width=0.48\linewidth, trim={0 0 25 32},clip]{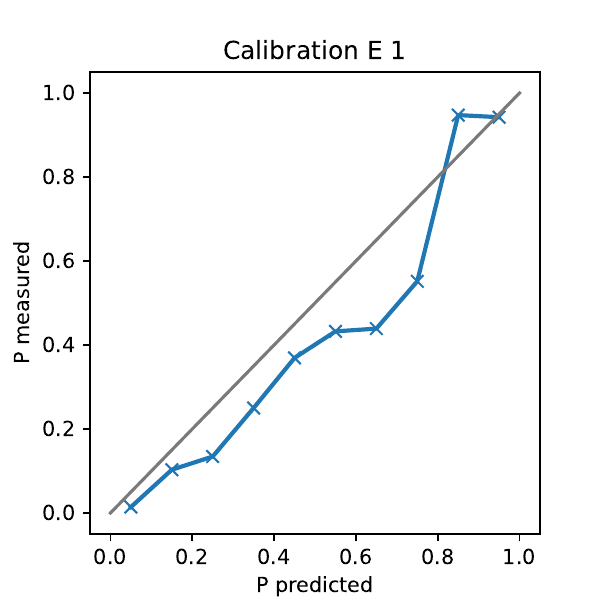}}
  \vspace{0.5mm}
  \caption{Evaluation of the calibration for simulated annealing in clustering scenarios with 3~clusters and 30/45~points respectively. All results generated with 1,000 problems in each scenario and 20,000 measurements for each clustering problem.}
  \label{fig:calibration_sim_num_points}
\end{figure}

\section{Cluster Calibration Evaluation}
\label{sec:clustering_performance}

This section extends the analysis of the calibration of our method to additional synthetic scenarios with more variation and increased problem size. The plots provided in this section are generated similarly to the main manuscript where first all clustering solutions $Z$ are accumulated in bins according to their estimated posterior probability $P(Z|\vct X)$, including all sampled but non-optimal solutions. After accumulation, the ratio of correct solutions in each bin is evaluated and plotted over the probability range of each bin. In the plots the diagonal represents the desired calibration.

Further calibration plots for the scenario with 3~clusters and an increasing number of total points are provided in Figure~\ref{fig:calibration_sim_num_points}. The scenarios are solved using simulated annealing with 20,000~measurements for each problem. The experiment with a total of 45~points in Figure~\ref{fig:calibration_3315} shows an overestimation of the posterior probability of the respective solutions. This can be attributed to two possible scenarios, where 1) the best solution is found, but not all relevant high-energy solutions are found during annealing and 2) the lowest-energy solution is not found and thus, the probability of all other solutions is overestimated. As the optimization problem becomes harder with an increasing number of points, the behavior is stronger in Figure~\ref{fig:calibration_3315} than in Figure~\ref{fig:calibration_3310}.

\section{Coreset Sparsification Performance}
\label{sec:coreset_sparsification_performance}
\begin{figure}[t!]
    \centering
    {\includegraphics[width=0.7\linewidth, clip]{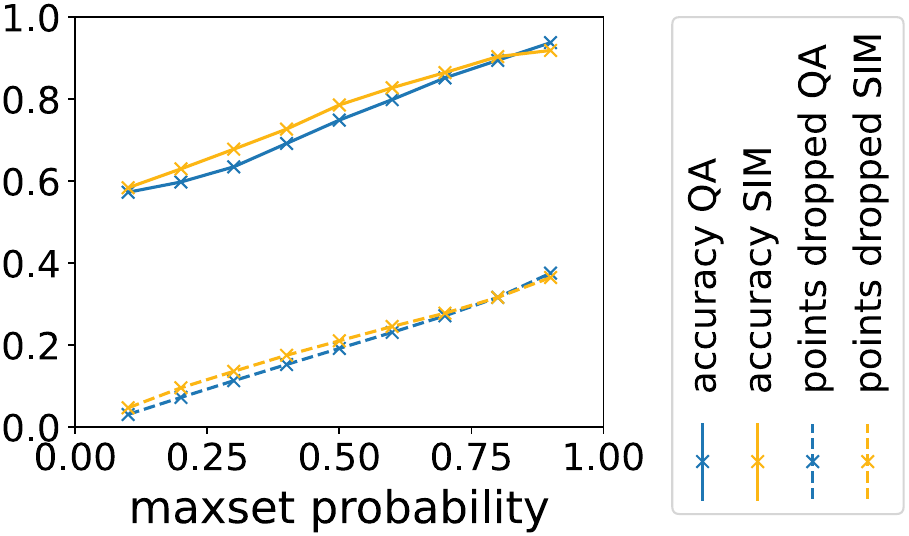}}
    \caption{Clustering accuracy wrt. removing uncertain points by merging coresets.}
    \label{fig:coreset_sparsification}
\end{figure}

The set of feasible solutions can be merged by using the calibrated confidence scores in Algorithm 1 introduced in the main manuscript. It sequentially removes uncertain points from the solution thus, increasing the solution probability. In Figure~\ref{fig:coreset_sparsification}, we show the probability of the best merged solution being correctly evaluated over the minimum solution probability of the sparsified coreset. It shows that our approach of removing single points can considerably increase the solution probability, thus highlighting the quality of the found coresets.

\section{Evaluation on IRIS}
\label{sec:performance_iris}
\begin{figure*}
    \vspace{-2mm}
    \centering
    \includegraphics[width=0.7\linewidth]{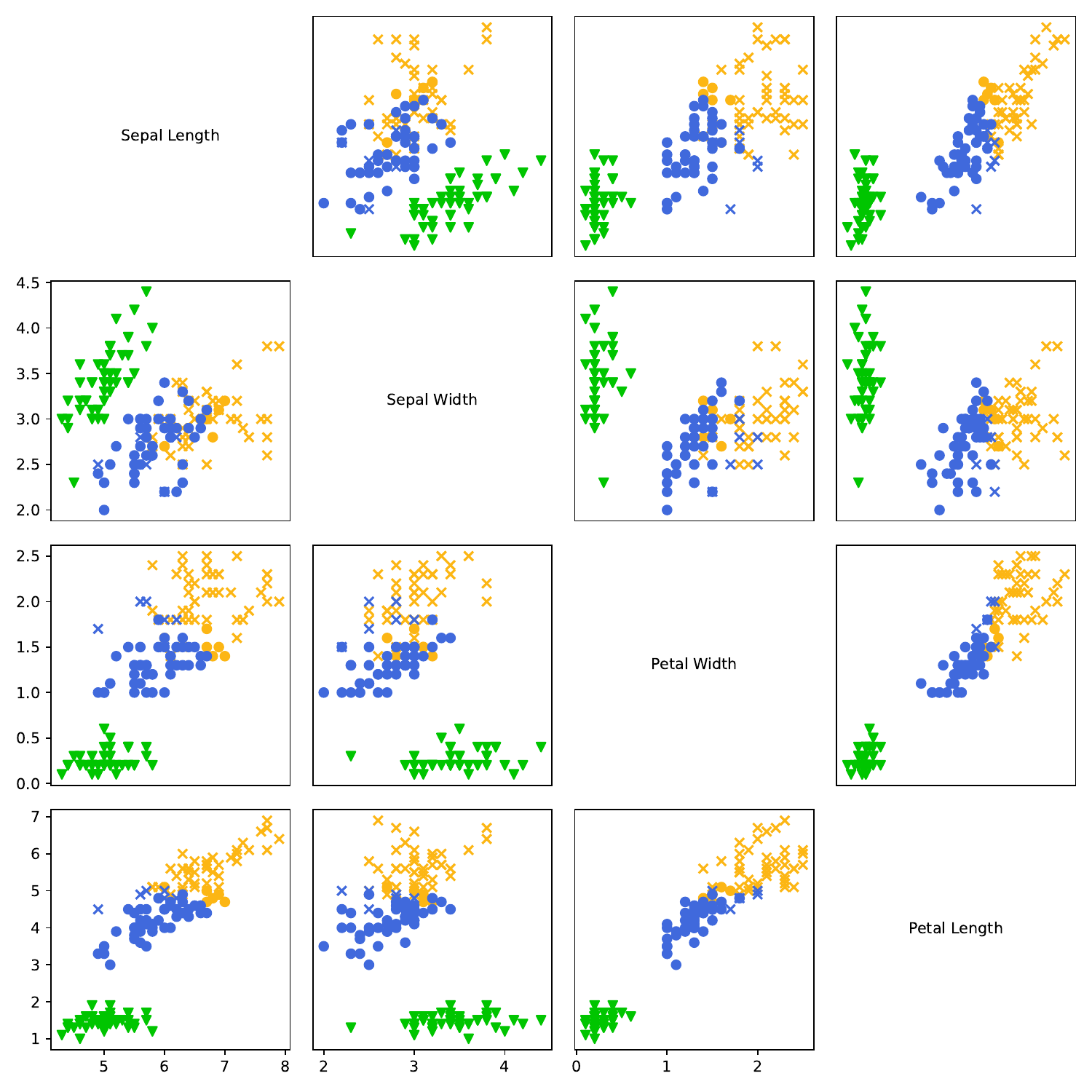}
    \caption{Clustering results on the IRIS dataset for simulated annealing and k-means.}
    \label{fig:IRIS_all_sim}
\end{figure*}

Table~2 in the main manuscript provides performance metrics for randomly subsampled versions of the IRIS dataset~\cite{fisher_use_1936}. In this section, we further evaluate the performance on the whole IRIS dataset, which contains 3 classes, 50 samples for each class and 4 features. We use simulated annealing with 20,000 measurements and balanced k-means to solve the IRIS clustering task, which both provide the same solution. The qualitative results are depicted in Figure~\ref{fig:IRIS_all_sim}, where all pairs of features are visualized. The shape of each sample represents the ground truth class and the color the result of the clustering algorithm. While the different feature pairs are plotted separately, the problem is solved as a single 4-dimensional clustering task. As results are identical with simulated annealing and balanced k-means, clustering metrics are also identical with a Completeness of 77.7\%, Adjusted Rand index 78.6\% and Fowlkes-Mallows Score of 85.6\%.

\section{Failure Cases}
\label{sec:failure_cases}
Analyzing the failure cases of our method provides valuable insight into the current state of quantum computing in computer vision, which aids to identify areas that need to be further investigated.

\subsection{K-means}
\begin{figure}[t]
  \centering
  \subfloat[\centering K-means.
    \label{fig:failure_kmeans_kmeans1}]{
    \includegraphics[width=0.48\linewidth, trim={20 30 20 30},clip]{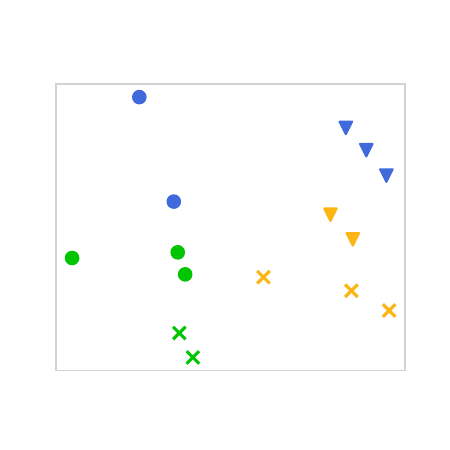}}
  \hfill
  \subfloat[\centering Simulated annealing.
    \label{fig:failure_kmeans_sim1}]{
    \includegraphics[width=0.48\linewidth, trim={20 30 20 30},clip]{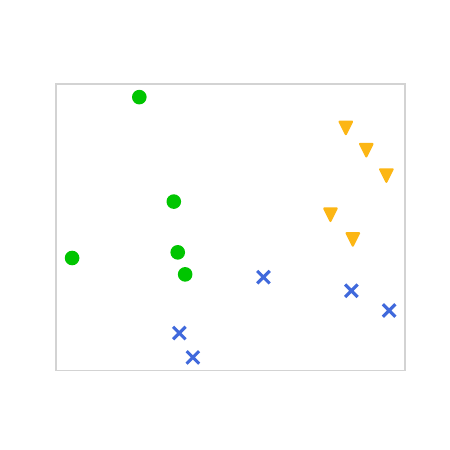}}

  \subfloat[\centering K-means.
    \label{fig:failure_kmeans_kmeans2}]{
    \includegraphics[width=0.48\linewidth, trim={20 48 20 48},clip]{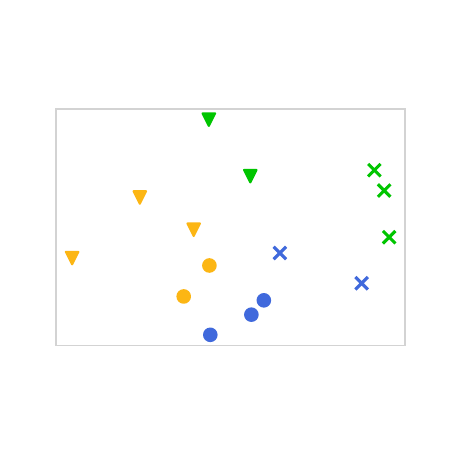}}
  \hfill
  \subfloat[\centering Simulated annealing.
    \label{fig:failure_kmeans_sim2}]{
    \includegraphics[width=0.48\linewidth, trim={20 48 20 48},clip]{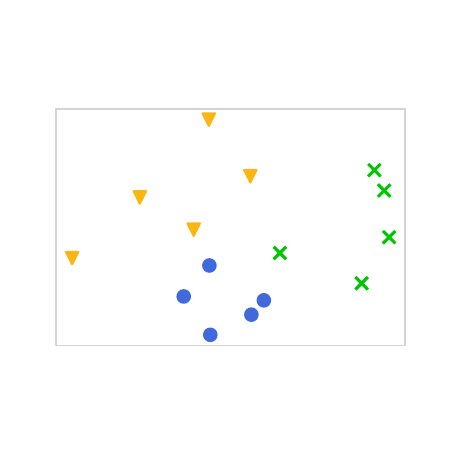}}
  \vspace{0.5mm}
  \caption{Failure cases for k-means clustering. While our formulation finds the correct solution, k-means returns a local minimum.}
  \label{fig:failure_kmeans}
\end{figure}

The analysis of synthetic problems in Table~1 in the main manuscript shows an advantage of our approach compared to the balanced k-means algorithm~\cite{malinen_balanced_2014} for the smaller clustering scenarios. These cases can be traced back to the k-means algorithm finding local minima where switching points given the last cluster means does not improve the data fit. This scenario is avoided in our formulation by jointly optimizing for the assignment and cluster centers. Two examples of such failure cases of k-means clustering are provided in Figure~\ref{fig:failure_kmeans}.

\subsubsection{Annealing based clustering}
\begin{figure}[t]
  \centering
    \subfloat[\centering Simulated annealing.
    \label{fig:failure_sim_sim1}]{
    \includegraphics[width=0.45\linewidth, trim={10 30 10 30},clip]{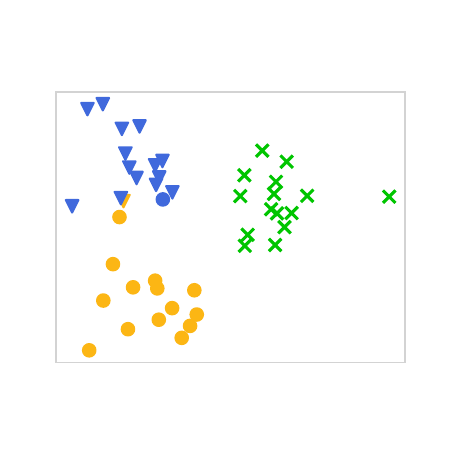}}
\hfill
  \subfloat[\centering K-means.
    \label{fig:failure_sim_kmeans1}]{
    \includegraphics[width=0.45\linewidth, trim={10 30 10 30},clip]{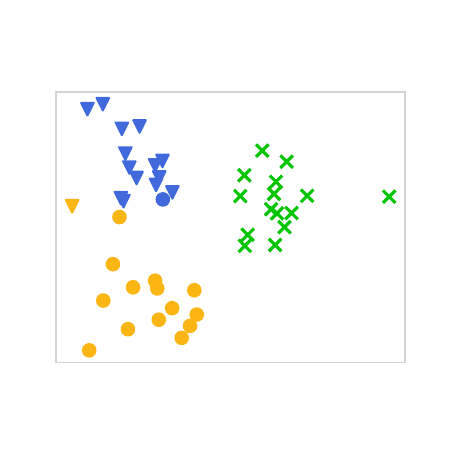}}

  \subfloat[\centering Simulated annealing.
    \label{fig:failure_sim_sim2}]{
    \includegraphics[width=0.45\linewidth, trim={10 25 10 25},clip]{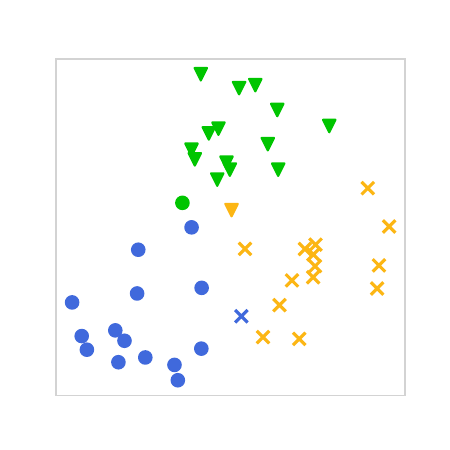}}
  \hfill
    \subfloat[\centering K-means.
    \label{fig:failure_sim_kmeans2}]{
    \includegraphics[width=0.45\linewidth, trim={10 25 10 25},clip]{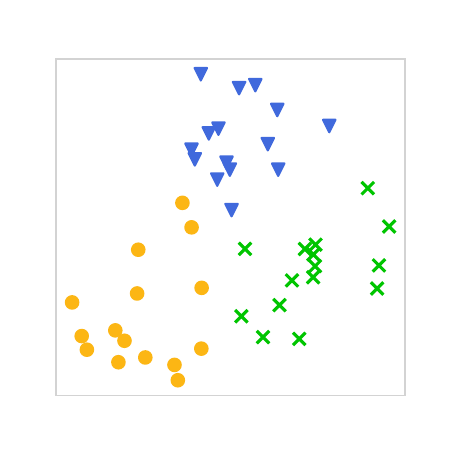}}

  \vspace{0.5mm}
  \caption{Failure cases for simulated annealing clustering.}
  \label{fig:failure_sim}
\end{figure}

The scenario with a total of 45 Points in 3 Clusters shows an advantage of k-means in the number of correctly solved problems compared to our approach using simulated annealing. The main source for this behavior lies in not finding the lowest energy and thus, the optimal solution of the clustering problem, as depicted in Figures~\ref{fig:failure_sim_sim1} and~\ref{fig:failure_sim_kmeans1}. Another source of error in this scenario is shown if Figures~\ref{fig:failure_sim_sim2} and~\ref{fig:failure_sim_kmeans2}, where the local k-means solution corresponds to the ground truth, even though it has a higher energy. As the solutions returned by our approach are still dense clusters, the clustering metrics remain competitive with the balanced k-means approach.

\section{Limitations}
\label{sec:limitations}
Our work aims at demonstrating the potential of using a quantum computer as a sampler for k-means clustering, in order to find multiple likely solutions and their associated calibrated posterior probabilities. Given the novelty of applying quantum computing to computer vision, it's natural that many works in this area, including ours, still come with limitations.

Current quantum computers are still limited in their fidelity of qubit couplings, which represent the terms of the quadratic cost function. This requires a careful selection of Lagrangian multipliers, which adds additional computational cost in the current formulation. With improving AQCs, this problem can be reduced and help to increase the problem size, as well as the robustness of the formulation. Another hardware limitation is the restricted connectivity between qubits. In the D-Wave Advantage 2 prototype used in this work, each qubit is coupled to up to 20 neighbors. This requires to build chains of qubits to represent a dense cost-matrix. Therefore, investigating sparse representations for clustering that reduce the required chain length can help to embed larger problems on the AQC.

Finally, our clustering approach is following the k-means cost function, with an identity covariance matrix. While this can model a range of practical problems, where the distribution of the data can directly be influenced, future work should investigate formulations of higher-order terms.

\addtolength{\textheight}{-14cm}

